\documentclass[runningheads]{llncs}

\usepackage{graphicx}
\usepackage{comment}
\usepackage{color}
\usepackage{url}

\newif\ifreview
\reviewfalse

\ifreview
	\usepackage{lineno}

	\linenumbers
\fi

\usepackage{cite}
\usepackage{times}
\usepackage{epsfig}
\usepackage{graphicx}
\usepackage{amsmath, amssymb}
\usepackage{authblk}
\usepackage{xspace}

\usepackage[pagebackref=true,breaklinks=true,letterpaper=true,colorlinks,bookmarks=false]{hyperref}

\usepackage{graphicx}
\usepackage{float} 
\usepackage{booktabs}  
\usepackage{color}  
\usepackage{amsmath,amssymb,amsthm}  
\usepackage{bm}  
\usepackage{times}  
\usepackage{lipsum} 
\usepackage{mathtools} 
\usepackage{placeins}  
\usepackage{fancyvrb} 
\usepackage{siunitx}  

\newcommand{\ra}[1]{\renewcommand{\arraystretch}{#1}}

\usepackage[super]{nth}
\usepackage{nicefrac}
\newcommand\figref[1]{Fig.~{\ref{#1}}}
\newcommand\tabref[1]{Table~{\ref{#1}}}

\newcommand{\norm}[1]{\left\lVert #1 \right\rVert}  

\def\rot#1{\rotatebox{90}{#1}}  

\def\tt{\mathbf{t}}

\newcommand{\DatasetName}{\textit{HanCo}}

\makeatletter
\DeclareRobustCommand\onedot{\futurelet\@let@token\@onedot}
\def\@onedot{\ifx\@let@token.\else.\null\fi\xspace}

\def\eg{\emph{e.g}\onedot}

\def\etal{\emph{et al}\onedot}
\makeatother
\begin{document}


\def\SubNumber{067}

\def\GCPRTrack{Regular Track}

\title{Contrastive  Representation Learning for Hand Shape Estimation}

\ifreview
\else

	\author{Christian Zimmermann\inst{*} \and Max Argus\inst{*} \and Thomas Brox}
	\institute{University of Freiburg, Germany\\
	* These authors contributed equally\\
	\vspace{.5em}\url{https://lmb.informatik.uni-freiburg.de/projects/contra-hand/}}
\fi

\maketitle              

    \begin{abstract}
    This work presents improvements in monocular hand shape estimation by building on top of recent advances in unsupervised learning. We extend momentum contrastive learning and contribute a structured collection of hand images, well suited for visual representation learning, which we call \DatasetName{}. We find that the representation learned by established contrastive learning methods can be improved significantly by exploiting advanced background removal techniques and multi-view information. These allow us to generate more diverse instance pairs than those obtained by augmentations commonly used in exemplar based approaches. Our method leads to a more suitable representation for the hand shape estimation task and shows a $4.7$\% reduction in mesh error and a $3.6$\% improvement in F-score compared to an ImageNet pretrained baseline. We make our benchmark dataset publicly available, to encourage further research into this direction.
    

\end{abstract}

    \section{Introduction}
Leveraging unlabeled data for training machine learning is a long standing goal in research and its importance has increased dramatically with the advances made by data-driven deep learning methods.
Using unlabeled data is an attractive proposition, because more training data usually leads to improved results. On the other hand, label acquisition for supervised training is difficult, time-consuming, and cost intensive.

While the use of unsupervised learning is conceptually desirable the research community struggled long to compete with simple transfer learning approaches using large image classification benchmark datasets. For a long time, there has been a substantial gap between the performance of these methods and the results of supervised pretraining on ImageNet. However, recent work \cite{DBLP:conf/cvpr/He0WXG20} succeeded to surpass ImageNet based pretraining for multiple downstream tasks. The core innovation was to use a consistent, dynamic and large dictionary of embeddings in combination with a contrastive loss, which is a practice we are following in this work as well. Similarly, we make use of strong geometric and color space augmentations, like flipping, cropping, as well as modification of hue and brightness, to generate positive pairs at training time. 

Additionally, we find that large improvements lie in the use of strategies that extend the simple exemplar strategy of using a single image and heavy augmentation to generate a positive pair of samples. More precisely, we explore sampling strategies that exploit the structure in \DatasetName{} that is available without additional labeling cost. The data was captured in an controlled multi-view setting as video sequences, which allows us to easily extract foreground segmentation masks and sample correlated hand poses by selecting simultaneously recorded images of neighboring cameras. This allows us to generate more expressive positive pairs during self-supervised learning.


\begin{figure}[!t]
\centering
 \centering
 \vspace{2.5em}
 \hspace{-3.5mm}\includegraphics[width=.85\linewidth]{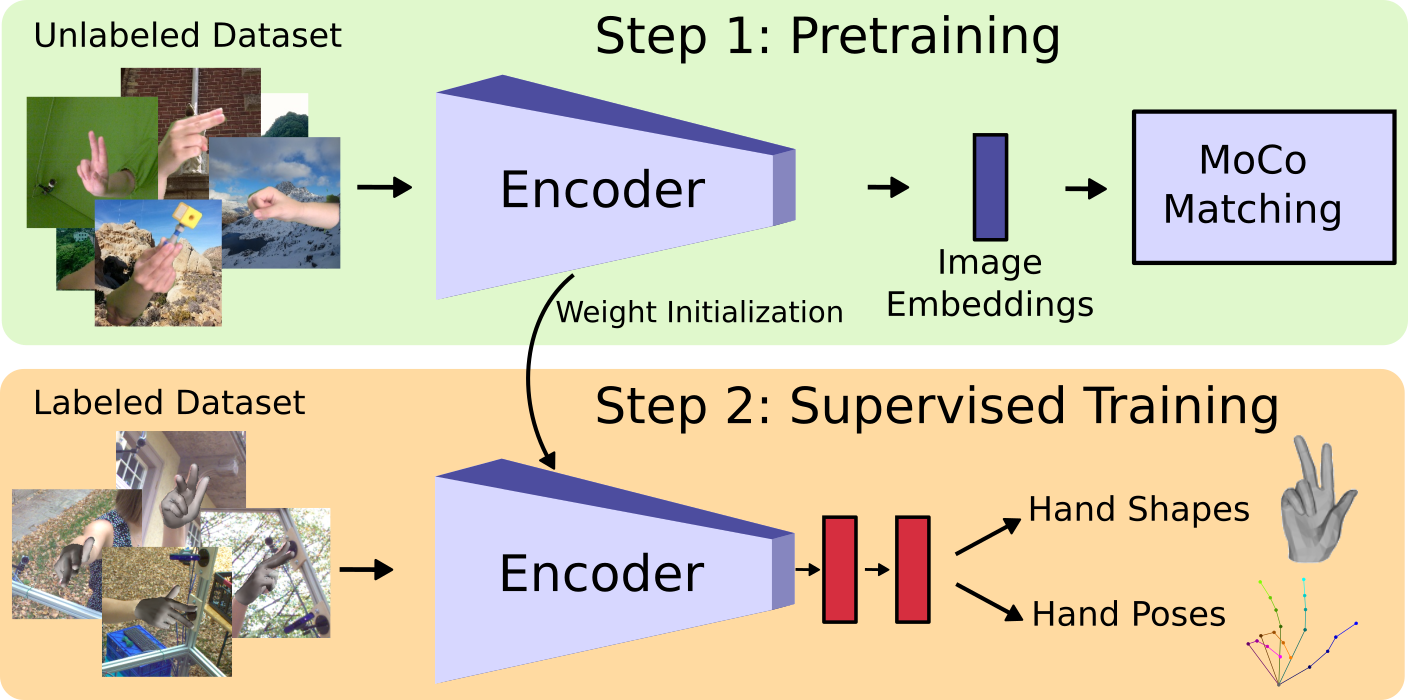}
 \caption{We train a self-supervised feature representation on our proposed large dataset of unlabeled hand images. The resulting encoder weights are then used to initialize supervised training based on a smaller labeled dataset. This pretraining scheme yields useful image embeddings that can be used to query the dataset, as well as increasing the performance of hand shape estimation.
 }
 \label{fig:teaser}
\vspace{-0.2cm}
\end{figure}

Hand shape estimation is a task where it is inherently hard to acquire diverse training data at a large scale. This stems from frequent ambiguities and its high dimensionality, which further raises the value of self-supervision techniques. Its need for large amounts of training data also makes hand shape estimation an ideal testbed for developing self-supervision techniques. Most concurrent work in the field of hand shape estimation follows the strategy of weak-supervision, where other modalities are used to supervise hand shape training indirectly.
Instead, we explore an orthogonal approach: pretraining the network on data of the source domain which eliminates both the need for hand shape labels as well as additional modalities.

In our work, we find that using momentum contrastive learning yields a valuable starting point for hand shape estimation. It can find a meaningful visual representation from pure self-supervision that allows us to surpass the ImageNet pretrained baseline significantly. We provide a comprehensive analysis on the learned representation and show how the procedure can be used for identifying clusters of hand poses within the data or perform retrieval of similar poses. 

For the purpose of further research into self-supervised pretraining we release a dataset that is well structured for this purpose it includes a) a larger number of images, b) temporal structure, and c) multi-view camera calibration.

\begin{figure*}[!t]
    \centering
    \includegraphics[width=1.0\linewidth, height=0.3\linewidth]{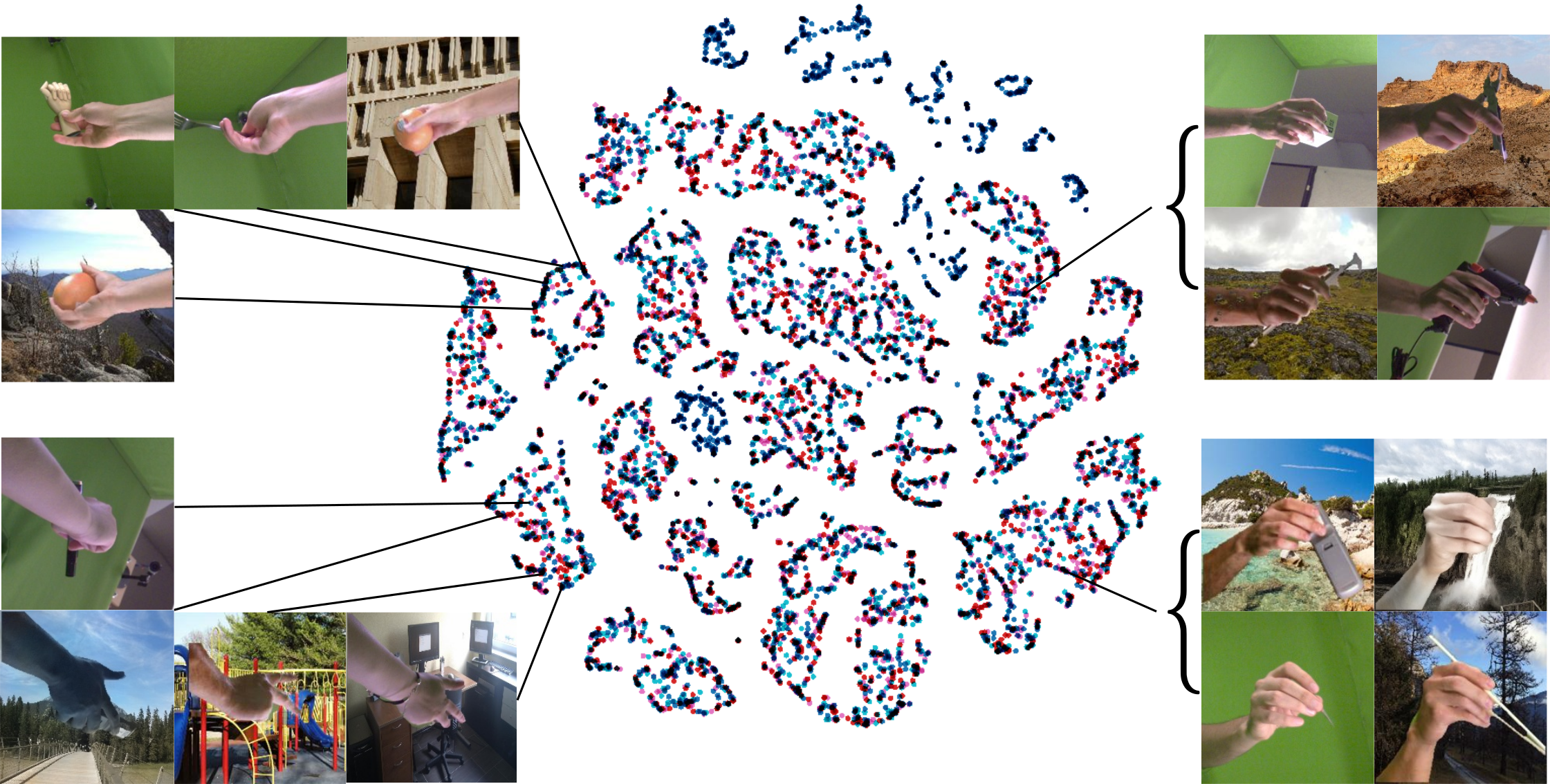}
    \caption{Shown is a two dimensional \textit{t-SNE} embedding of the image representation found by unsupervised visual representation learning. On the left hand side, we show that similar hand poses are located in proximity to each other. On the right hand side examples of nearly identical hand poses are shown which are approximately mapped to the same point. The color of the dots indicates which pre-processing method has been applied to the sample, a good mixing shows that the embedding focuses on the hand pose instead. Blue represents the unaltered camera-recorded images, red are images that follow the simple cut and paste strategy, while pink and bright blue correspond to images processed with the methods by Tsai \etal\cite{DBLP:conf/cvpr/TsaiSLSL017} and Zhang \etal\cite{DBLP:journals/tog/ZhangZIGLYE17} respectively. 
    }
    \label{fig:method}
\end{figure*}

\section{Related work}

\noindent\textbf{Visual Representation Learning}
Approaches that aim to learn representations from collection of images without any labels can be roughly categorized into generative and discriminative approaches. While earlier work was targetting generative approaches the focus shifted towards discriminative methods that either leverage contrastive learning or formulate auxiliary tasks using pseudo labels.

Generative approaches aim to recover the input, subject to a certain type of pertubation or regularization. Examples are DCGAN by Radford \etal\cite{DBLP:journals/corr/RadfordMC15}, image colorization \cite{DBLP:conf/eccv/ZhangIE16}, denoising autoencoders \cite{DBLP:conf/icml/VincentLBM08} or image in-painting \cite{DBLP:conf/cvpr/PathakKDDE16}.

Popular auxiliary tasks include solving Jigsaw Puzzles \cite{DBLP:conf/eccv/NorooziF16},
predicting image rotations \cite{DBLP:conf/iclr/GidarisSK18}, or clustering features during training \cite{DBLP:conf/eccv/CaronBJD18}. 

In contrast to auxiliary tasks the scheme of contrastive loss functions \cite{DBLP:conf/cvpr/HadsellCL06}, doesn't define pseudo-labels, but uses a dynamic dictionary of keys that are sampled from the data and resemble the current representation of samples via the encoding network. For training one matching pair of keys is generated and the objective drives the matching keys to be similar, while being dissimilar to the other entries in the dictionary.
Most popular downstream task is image classification, where contrastive learning approaches have yielded impressive results \cite{Caron2020, SimCLR2020, DBLP:journals/corr/abs-2003-04297}. In this setting pretrained features are evaluated by a common protocol in which these features are frozen and only a supervised linear classifier is trained on the global average pooling of these features.
In this work we follow the paradigm of contrastive learning, hence the recent successes, and build on the work by Chen \etal\cite{DBLP:journals/corr/abs-2003-04297}. However, we find retraining of the complete convolutional backbone is necessary for hand shape estimation, which is following the transfer learning by fine-tuning \cite{Donahue2014, Yosinski2014} idea. Furthermore, we extend sampling of matching keys beyond exemplar and augmentation-based strategies.

\begin{figure*}[!t]
\centering

\includegraphics[width=1.0\linewidth]{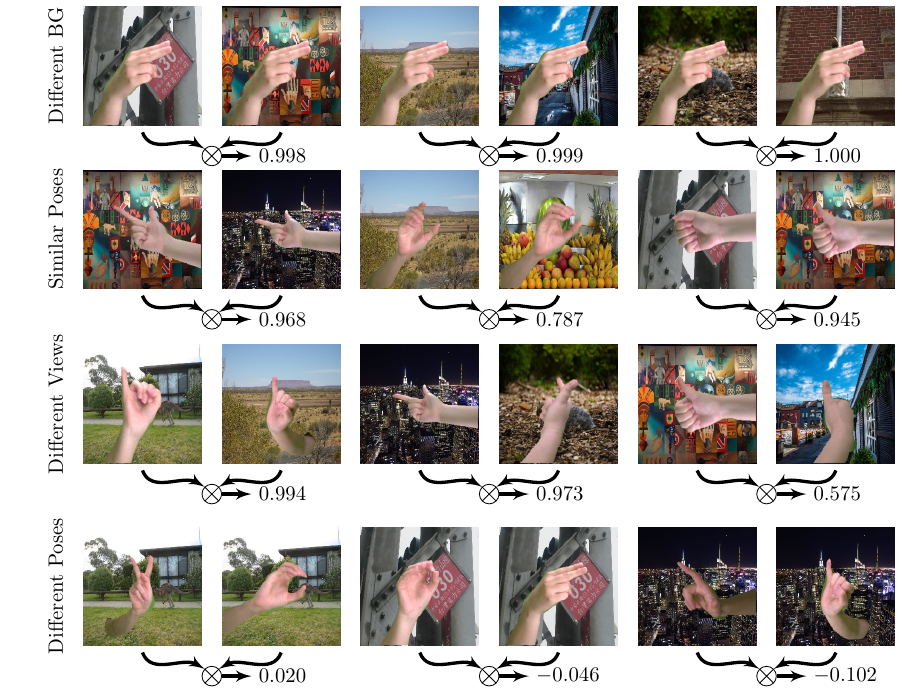}
    \caption{
    Examples showing the cosine-similarity scores for the embeddings produced by pretraining. 
    The first row shows embeddings for the same hand pose with different backgrounds, the embedding learns to focus on the hand and ignore the background.
    In the second row we show images for similar hand poses, with different background, these also score highly.
    In the third row different hand poses, with the same background produce low scores.}
    
\label{fig:embedd_qual}
\end{figure*}

\noindent\textbf{Hand Shape Estimation}
Most recent works in the realm of hand shape estimation rely on a deep neural network estimating parameters of a hand shape model. By far, the most popular choice is MANO presented by Romero \etal\cite{DBLP:journals/tog/0002TB17}. There are little approaches using another surface topology; examples are Moon \etal\cite{DBLP:conf/eccv/MoonSL20} and Ge \etal\cite{DBLP:conf/cvpr/GeRLXWCY19}, but MANO was used by the majority of works \cite{DBLP:conf/cvpr/BoukhaymaBT19, DBLP:conf/cvpr/HassonVTKBLS19, DBLP:conf/cvpr/KulonGKBZ20, DBLP:conf/iccv/ZimmermannCYRAB19} and is also used in our work. 

Commonly, these approaches are trained with a combination of losses, frequently incorporating additional supervision coming from shape derived modalities like depth \cite{DBLP:conf/eccv/CaiGCY18, DBLP:conf/eccv/MoonSL20, DBLP:conf/cvpr/WanPGY19}, silhouette \cite{DBLP:conf/cvpr/BaekKK19, DBLP:conf/cvpr/BoukhaymaBT19, DBLP:journals/sensors/MalikES19}, or keypoints \cite{DBLP:conf/cvpr/BoukhaymaBT19, DBLP:conf/cvpr/HassonVTKBLS19, DBLP:conf/cvpr/KulonGKBZ20, DBLP:conf/eccv/MoonSL20, DBLP:conf/iccv/ZimmermannCYRAB19}, which is referred to as weak-supervision. 
This allows to incorporate datasets into training that don't contain shape annotation, significantly reducing the effort for label acquisition. Sometimes, these approaches incorporate adversarial losses \cite{DBLP:conf/cvpr/KanazawaBJM18}, which can help to avoid degenerate solutions on the weakly labeled datasets.
Other works focusing on hand pose estimation propose weak supervision by incorporating biomechanical constraints \cite{DBLP:conf/eccv/SpurrIMHK20} or sharing a latent embedding space between multiple modalities \cite{DBLP:conf/cvpr/Spurr0PH18, DBLP:conf/cvpr/TheodoridisCSDD20}.

One specific way of supervision between modalities is leveraging geometric constraints of multiple camera views, which was explored in human pose estimation before: Rhodin \etal\cite{DBLP:conf/cvpr/RhodinSKCMMSF18} proposed to run separate 3D pose detectors per view and constrain the estimated poses with respect to each other. Simon \etal\cite{DBLP:conf/cvpr/SimonJMS17} presents an iterative bootstrapping procedure of self-labeling with using 3D consistency as a quality criterion. Yao \etal\cite{DBLP:conf/iccv/YaoJP19} and He \etal\cite{DBLP:conf/cvpr/HeYFY20a} supervise 2D keypoint detectors by explicitly incorporating epipolar constraints between two views.

In our work the focus is not on incorporating constraints by adding weak-supervision to the models' training, but we focus on finding a good initialization for supervised hand shape training using methods from unsupervised learning.

    \section{Approach}
\label{sec:approach}
Our approach to improve monocular hand shape estimation consists of two steps and is summarized in \figref{fig:teaser}: First, we are pretraining the CNN encoder backbone on large amounts of unlabeled data using unsupervised learning on a pretext task. Second, the CNN is trained for hand shape estimation in a supervised manner, using the network weights from the pretext task as initialization.

\noindent\textbf{Momentum Contrastive Learning}
\textit{MoCo} \cite{DBLP:conf/cvpr/He0WXG20} is a recent self-supervised learning method that performs contrastive learning as a dictionary look-up. \textit{MoCo} uses two encoder networks to encode different augmentations of the same image instance as query and key pairs.
Given two images $\bm{I}_i \in \mathbb{R}^{H \times W}$ and $\bm{I}_j \in \mathbb{R}^{H \times W}$ the embeddings are calculated as 
\begin{align}
    q &= f(\theta,~\bm{I}_i)    \text{\hspace{2em}and}\\
    k &= f(\tilde{\theta},~\bm{I}_j) \text{\hspace{2em}.}
\end{align}
which yields a query $q$ and key $k$. The same function $f$ is used in both cases, but parameterized differently. The query function uses $\theta$ which is directly updated by the optimization, while the key function $\tilde{\theta}$ is updated indirectly. At a given optimizations step $n$ it is calculated as 
\begin{align}
    \tilde{\theta}_{n} = m \cdot \tilde{\theta}_{n-1} + (1-m) \cdot \theta_{n}
\end{align}
using the momentum factor $m$, which is chosen close to $1.0$ to ensure a slowly adapting encoding of the key values $k$.

During training a large queue of dictionary keys $k$ is accumulated over iterations which allows for efficient training as a large set of negative samples has been found to be critical for contrastive training \cite{DBLP:conf/cvpr/He0WXG20}. Following this methodology \textit{MoCo} produces feature representations that transfer well to a variety of downstream tasks.
As training objective the InfoNCE loss \cite{DBLP:journals/corr/abs-1807-03748} is used, which relates the inner product of the matching key-query-pair to the inner products of all negative pairs in a softmax cross-entropy kind of fashion.

At test time the similarity of a key-value-pair can be computed using cosine similarity
\begin{align}
    \text{cossim}(q, ~k) = \frac{q \cdot k}{\norm{q}_2 \cdot \norm{k}_2}
    \label{eq:cossim}
\end{align}
which can return values ranging from $1.0$ in the similar case to $-1.0$ in the dissimilar case.

\textit{MoCo} relies entirely on standard image space augmentations to generate positive pairs of image during representation learning. A function $g(.)$, subject to a random vector $\zeta$, is applied to the same image instance $\bm{I}$ two times to generate different augmentations
\begin{align}
    \bm{I}_i = g(\bm{I},~\zeta_1) \\
    \bm{I}_j = g(\bm{I},~\zeta_2) 
\end{align}
that are considered as the matching pair. The function $g(.)$ performs randomized: crops of the image, color jitter, grayscale, and conversion and Gaussian blur. We omit randomized image flipping as this augmentation changes the semantic information of the hand pose.

The structured nature of \DatasetName{} allows us going beyond these augmentation-based strategies.
Here we are looking for strategies that preserve the hand pose, but change other aspects of the image. We investigate three different configurations a) background randomization, b) temporal sampling and c) multi-view sampling.

\DatasetName{} consists of short video clips that are recorded by multiple cameras simultaneously at \SI{5}{Hz}. The clips are up to \SI{40}{\sec} long and have an average length of \SI{14}{\sec}. For simplicity and without loss of generality we describe the sampling methods for a single sequence only. Extending the approach towards multiple sequences is straight forward, by first sampling a sequence and then applying the described procedure.
Formally, we can sample from a pool of images $\bm{I}_t^c$ recorded at a timestep $t$ from camera $c$.

During background randomization we use a single image $\bm{I}_t^c$ with its foreground segmentation as source to cut the hand from the source image and paste it into a randomly sampled background image.  Example outputs of background randomization are shown in the first row of \figref{fig:embedd_qual}.

For temporal sampling we exploit the fact, that our unlabeled dataset stems from a video stream which naturally constraints subsequent hand poses to be highly correlated. A positive pair of samples is generated by sampling two neighboring frames $\bm{I}_t^c$ and $\bm{I}_{t+1}^c$ for a given camera $c$. Due to hand movement the hand posture is likely to change slightly from $t$ to $t+1$, which naturally captures the fact that similar poses should be encoded with similar embeddings. 

As the data is recorded using a calibrated multi-camera setup and frame capturing is temporally synchronized, views from different cameras at a particular point in time show the same hand pose. This can be used as a powerful method of "augmentation" as different views change many aspects of the image but not the hand pose. Consequently, we generate a positive sample pair $\bm{I}_t^{c_1}$ and $\bm{I}_{t}^{c_2}$ in the multi-view case by sampling neighboring cameras $c_1$ and $c_2$ at a certain timestep $t$. The dataset contains an $8$ camera setup, with cameras mounted on each of the corners of a cube, in order to simplify the task of instance recognition, we chose to sample neighboring cameras, meaning those connected by no more than one edge of the cubical fixture.

\begin{figure}[!tb]
    \centering

\newcommand{\QiW}{0.16}

\begin{tabular}{@{}c@{\hspace{5mm}}c@{\hspace{1mm}}c@{\hspace{1mm}}c@{\hspace{1mm}}c@{}}
    Query      & \multicolumn{3}{c}{Next most similar}\\
    Image      & First & Second & Third \\
    \includegraphics[width=\QiW\linewidth]{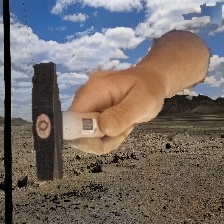} &
    \includegraphics[width=\QiW\linewidth]{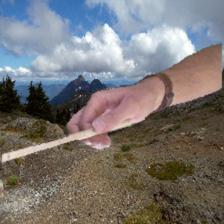} &
    \includegraphics[width=\QiW\linewidth]{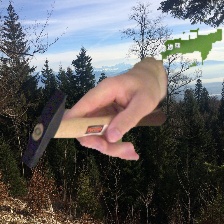} &
    \includegraphics[width=\QiW\linewidth]{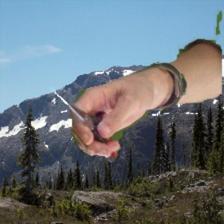} \\
    
    \includegraphics[width=\QiW\linewidth]{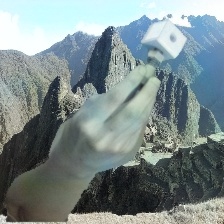} &
    \includegraphics[width=\QiW\linewidth]{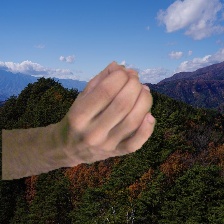} &
    \includegraphics[width=\QiW\linewidth]{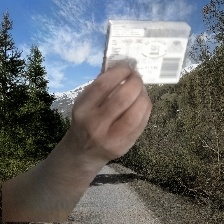} &
    \includegraphics[width=\QiW\linewidth]{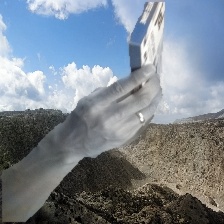} \\
    
    \includegraphics[width=\QiW\linewidth]{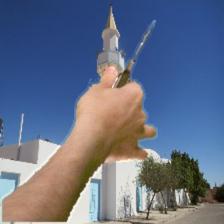} &
    \includegraphics[width=\QiW\linewidth]{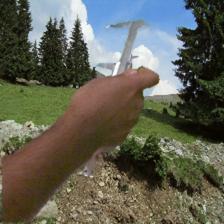} &
    \includegraphics[width=\QiW\linewidth]{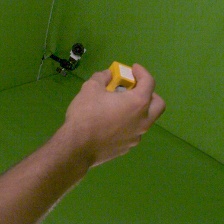} &
    \includegraphics[width=\QiW\linewidth]{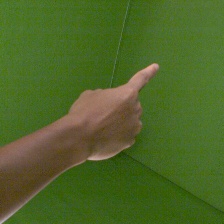} \\
    
    \includegraphics[width=\QiW\linewidth]{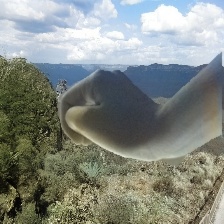} &
    \includegraphics[width=\QiW\linewidth]{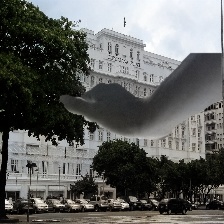} &
    \includegraphics[width=\QiW\linewidth]{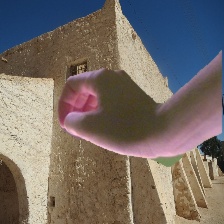} &
    \includegraphics[width=\QiW\linewidth]{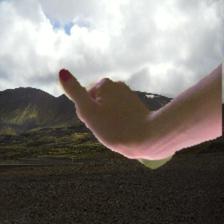}
\end{tabular}

    \caption{Given a query image our learned embedding allows to identify images showing similar poses. This enables identifying clusters in the data, without the need of pose annotations. The nearest neighbors are queried from a random subset of 25,000 samples of \DatasetName{}.}
    \label{fig:retrieval}
\end{figure}

\noindent\textbf{Hand Shape Estimation}
Compared to unsupervised pretraining the network architecture used for shape estimation is modified by changing the number of neurons of the last fully-connected layer from $128$ to $61$ neurons in order to estimate the \textit{MANO} parameter vector. Consequently, the approach is identical to the one presented by Zimmermann \etal\cite{DBLP:conf/iccv/ZimmermannCYRAB19} with the difference being only that the weights of the convolutional backbone are initialized through unsupervised contrastive learning and not ImageNet pretraining.
The network is being trained to estimate the \textit{MANO} parameter vector $\bm{\tilde{\theta}} \in \mathbb{R}^{61}$ using the following loss:
\begin{align}
    \mathcal{L} = w_\text{3D} &\norm{\bm{P} - \tilde{\bm{P}}} + \nonumber \\  w_\text{2D} &\norm{\Pi (\bm{P}) - \Pi ( \tilde{\bm{P}} )} +   \nonumber \\
    w_\text{p} &\norm{\bm{\theta} - \tilde{\bm{\theta}}} \text{.}
\end{align}

We deploy $L_2$ losses for all components and weight with $w_\text{3D} = 1000$, $w_\text{2D} = 10$, and $w_\text{p} = 1$ respectively. To derive the predicted keypoints $\bm{\tilde{P}}$ from the estimated shape $\bm{\tilde{\theta}}$ in a differentiable way the MANO \etal\cite{DBLP:journals/tog/0002TB17} model implementation in \textit{PyTorch} by Hasson \etal\cite{DBLP:conf/cvpr/HassonVTKBLS19} is used. $\bm{P} \in \mathbb{R}^{21}$ is the ground truth 3D location of the hand joints and $\bm{\theta}$ the ground truth set of MANO parameter, both of which are provided by the training dataset. Denoted by $\Pi (.)$ is the projection operator mapping from 3D space to image pixel coordinates.

\begin{figure*}[!ht]
    \centering
    \newcommand{\figH}{0.075}
    \newcommand{\figW}{0.075}

\begin{tabular}{@{}c@{\hspace{1mm}}c@{\hspace{1mm}}c@{\hspace{1mm}}c@{\hspace{1mm}}c@{\hspace{1mm}}c@{\hspace{1mm}}c@{\hspace{1mm}}c@{\hspace{1mm}}c@{\hspace{1mm}}c@{\hspace{1mm}}c@{\hspace{1mm}}c@{\hspace{1mm}}c@{\hspace{1mm}}c@{}}
     \raisebox{.2\height}{\rot{\small RGB}} 
        & \includegraphics[height=\figH\linewidth,width=\figW\linewidth]{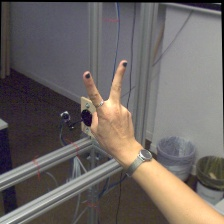}
        & \includegraphics[height=\figH\linewidth,width=\figW\linewidth]{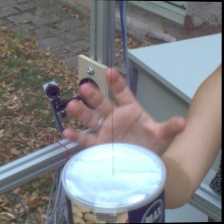}
        & \includegraphics[height=\figH\linewidth,width=\figW\linewidth]{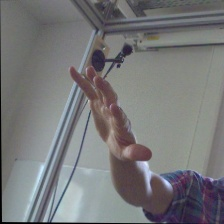}
        & \includegraphics[height=\figH\linewidth,width=\figW\linewidth]{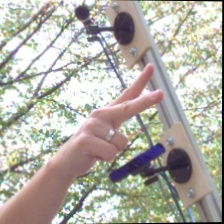}
        & \includegraphics[height=\figH\linewidth,width=\figW\linewidth]{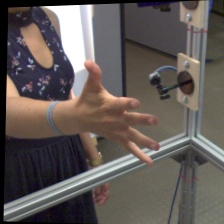}
        & \includegraphics[height=\figH\linewidth,width=\figW\linewidth]{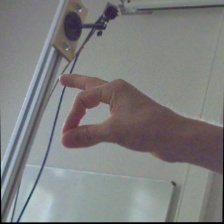}
        & \includegraphics[height=\figH\linewidth,width=\figW\linewidth]{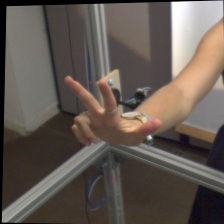}
        & \includegraphics[height=\figH\linewidth,width=\figW\linewidth]{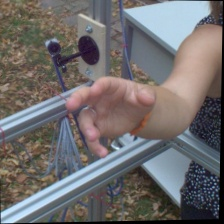}
        & \includegraphics[height=\figH\linewidth,width=\figW\linewidth]{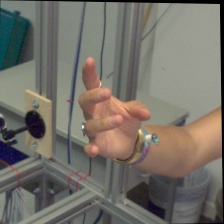}
        & \includegraphics[height=\figH\linewidth,width=\figW\linewidth]{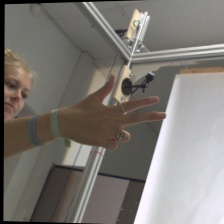}
        & \includegraphics[height=\figH\linewidth,width=\figW\linewidth]{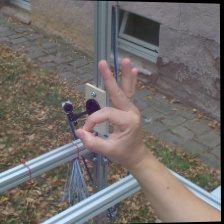}
        & \includegraphics[height=\figH\linewidth,width=\figW\linewidth]{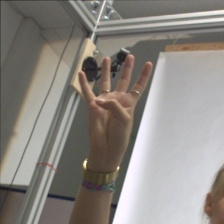} \\
     \raisebox{.2\height}{\rot{\small Ours}}
        & \includegraphics[height=\figH\linewidth,width=\figW\linewidth]{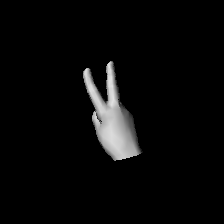}
        & \includegraphics[height=\figH\linewidth,width=\figW\linewidth]{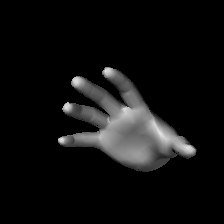}
        & \includegraphics[height=\figH\linewidth,width=\figW\linewidth]{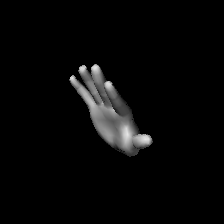}
        & \includegraphics[height=\figH\linewidth,width=\figW\linewidth]{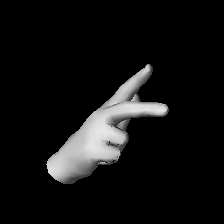}
        & \includegraphics[height=\figH\linewidth,width=\figW\linewidth]{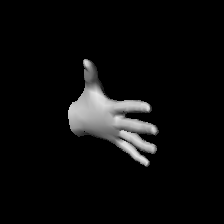}
        & \includegraphics[height=\figH\linewidth,width=\figW\linewidth]{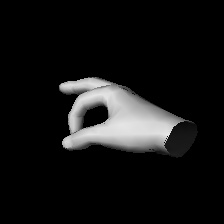}
        & \includegraphics[height=\figH\linewidth,width=\figW\linewidth]{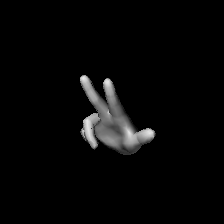}
        & \includegraphics[height=\figH\linewidth,width=\figW\linewidth]{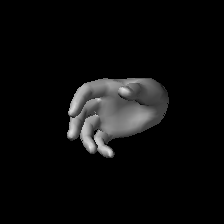}
        & \includegraphics[height=\figH\linewidth,width=\figW\linewidth]{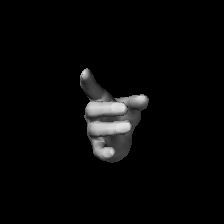}
        & \includegraphics[height=\figH\linewidth,width=\figW\linewidth]{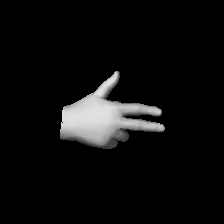}
        & \includegraphics[height=\figH\linewidth,width=\figW\linewidth]{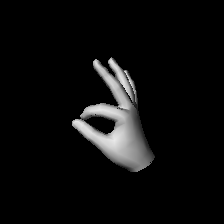}
        & \includegraphics[height=\figH\linewidth,width=\figW\linewidth]{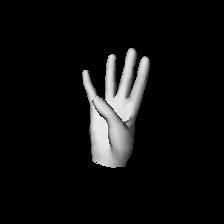} \\
     \raisebox{0.2\height}{\rot{\small \cite{DBLP:conf/iccv/ZimmermannCYRAB19}}}
        & \includegraphics[height=\figH\linewidth,width=\figW\linewidth]{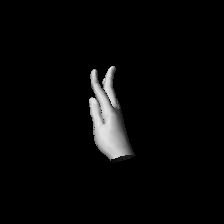}
        & \includegraphics[height=\figH\linewidth,width=\figW\linewidth]{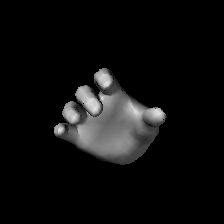}
        & \includegraphics[height=\figH\linewidth,width=\figW\linewidth]{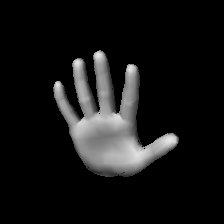}
        & \includegraphics[height=\figH\linewidth,width=\figW\linewidth]{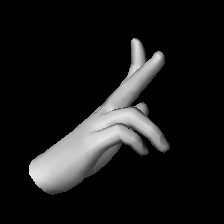}
        & \includegraphics[height=\figH\linewidth,width=\figW\linewidth]{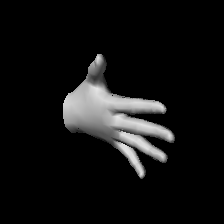}
        & \includegraphics[height=\figH\linewidth,width=\figW\linewidth]{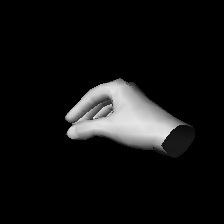}
        & \includegraphics[height=\figH\linewidth,width=\figW\linewidth]{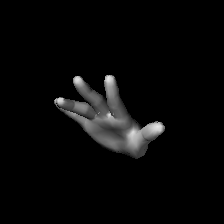}
        & \includegraphics[height=\figH\linewidth,width=\figW\linewidth]{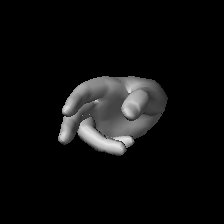}
        & \includegraphics[height=\figH\linewidth,width=\figW\linewidth]{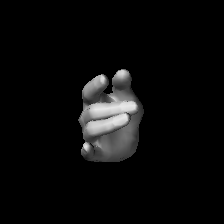}
        & \includegraphics[height=\figH\linewidth,width=\figW\linewidth]{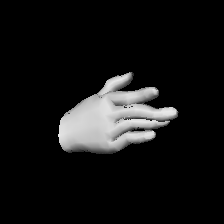}
        & \includegraphics[height=\figH\linewidth,width=\figW\linewidth]{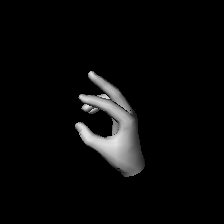}
        & \includegraphics[height=\figH\linewidth,width=\figW\linewidth]{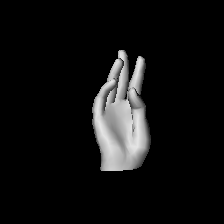} \\
\end{tabular}

    \caption{Qualitative comparison of MANO predictions between \textit{Ours-Multi view} and the approach by Zimmermann \etal\cite{DBLP:conf/iccv/ZimmermannCYRAB19} showing improvements in hand mesh predictions yielded by our self-supervised pretraining on the evaluation split of \textit{FreiHAND} \cite{DBLP:conf/iccv/ZimmermannCYRAB19}. Generally, our predictions look seem to capture the global pose and grasp of the hand more accurately, which results into a lower mesh error and higher F@5 score. }
    \label{fig:qual_results}
    \vspace{-2ex}
\end{figure*}

\section{Experiments}
\label{sec:experiments}

\noindent\textbf{Dataset}
Our experiments are conducted on data recorded by Zimmermann~\etal\cite{DBLP:conf/iccv/ZimmermannCYRAB19}, which the authors kindly made available to us. 
The images show 32 subjects which are recorded by 8 cameras simultaneously that are mounted on a cubical aluminum fixture. The cameras face towards the center of the cube and are calibrated. One part of the data was recorded against a green background, which allows extracting of the foreground segmentation automatically and to perform background randomization without any additional effort. Another part of the data was intended for evaluation and was recorded without green backgrounds.
For a subset of both parts there are hand shape labels, which were created by the annotation method \cite{DBLP:conf/iccv/ZimmermannCYRAB19}. The set of annotated frames was published before as \textit{FreiHAND} \cite{DBLP:conf/iccv/ZimmermannCYRAB19}, which we use for evaluation of the supervised hand shape estimation approaches. 

For compositing the hand foreground with random backgrounds we have collected $2193$ background images from Flickr, that are showing various landscapes, city scenes and indoor shots. We manually inspect these images, to ensure they don't contain shots targeted at humans.
There are three different methods to augment the cut and paste version: The colorization approach by Zhang \etal\cite{DBLP:journals/tog/ZhangZIGLYE17} is used in both its automatic and sampling-based mode. Also we use the deep harmonization approach proposed by Tsai \etal\cite{DBLP:conf/cvpr/TsaiSLSL017} that can remove color bleeding at the foreground boundaries. The background post-processing methods are also reflected by the point colors in the \textit{t-SNE} embedding plot (\figref{fig:method}).

The complete dataset is used for visual representation learning and we refer to it as \DatasetName{}, which provides $107,538$ recorded time instances or poses, each recorded by $8$ cameras. All available frames are used for unsupervised training, which results into $860,304$ frames, while a subset of $63,864$ frames contains hand shape annotations and are recorded against green screen, which we use for supervised training of the monocular hand shape estimation network.

\noindent\textbf{Training details}
For training of neural networks the \textit{PyTorch} framework is used and we rely on ResNet-50 as convolutional backbone \cite{DBLP:conf/cvpr/HeZRS16, DBLP:journals/corr/abs-1903-10520}.

During unsupervised training we follow the procedure by Chen \etal\cite{DBLP:journals/corr/abs-2003-04297} and train with the following hyper-parameters: a base learning rate of $0.015$ is used, which is annealed following a cosine schedule over $100$ epochs. We train with a batch size of $128$ and an image size of $224\times224$ pixels. We follow the augmentations of Chen \etal\cite{DBLP:journals/corr/abs-2003-04297}, but skip image flipping.
For supervised training we follow the training schedule by Zimmermann \etal\cite{DBLP:conf/iccv/ZimmermannCYRAB19}. The network is trained for $500,000$ steps with a batch size of $16$. A base learning rate of $0.0001$ is used, and decayed by a factor of $0.5$ after $220,000$ and $340,000$ steps.

\noindent\textbf{Evaluation of Embeddings}
First, we perform a qualitative evaluation of the learned embedding produced by pretraining.
For this purpose, we sample a random subset of $25,000$ images from the unlabeled dataset. The query network is used to compute a $128$ dimensional feature vector for each image. Using the feature vectors we find a \textit{t-SNE} \cite{ljpvd2008visualizing} representation that is shown in \figref{fig:method}. It is apparent, that similar poses cluster closely together, while different poses are clearly separated.

Pairs of images, together with their cosine similarity scores \eqref{eq:cossim} are shown in \figref{fig:qual_results}, the cosine similarity scores reveal many desirable properties of the embedding: The representation is invariant to the picked background (first row), small changes in hand pose only result in negligible drop in similarity (second row). Viewing the same hand pose from different directions results in high similarity scores, though this can be subject to occlusion (third row). Large changes in hand pose induce a significant drop of the score (last row). 
This opens up the possibility to use the learned embedding for retrieval tasks, which is shown in \figref{fig:retrieval}. Given a query image it is possible to identify images showing similar and different hand poses, without an explicit hand pose annotation.

\begin{table*}[ht]
\newcolumntype{Z}{S[table-format=2.1,table-auto-round]}
\centering
\ra{1.05}
\small
\begin{tabular}{|c|c|c|c|c| }
\hline
Method & Mesh Error in \SI{}{cm} $~\downarrow$ & F@\SI{5}{mm} $~\uparrow$ & F@\SI{15}{mm} $~\uparrow$\\
\hline\hline
Boukhayma \etal\cite{DBLP:conf/cvpr/BoukhaymaBT19} & $1.30$ & $0.435$ & $0.898$ \\
Hasson \etal\cite{DBLP:conf/cvpr/HassonVTKBLS19} & $1.32$ & $0.436$ & $0.908$ \\
Zimmermann \etal\cite{DBLP:conf/iccv/ZimmermannCYRAB19} & $1.07$ & $0.529$ & $0.935$ \\
Ours-Fixed & $2.13$ & $0.299$ & $0.803$ \\
Ours-Fixed-BN & $1.25$ & $0.463$ & $0.914$ \\
Ours-Scratch & $1.24$ & $0.475$ & $0.915$ \\
Ours-Augmentation & $1.09$ & $0.521$ & $0.934$ \\
Ours-Background & $1.04$ & $0.538$ & $0.940$ \\
Ours-Temporal & $1.04$ & $0.538$ & $0.939$ \\
\textbf{Ours-Multi view} & $\pmb{1.02}$ & $\pmb{0.548}$ & $\pmb{0.943}$ \\
\hline
\end{tabular}
\vspace{1mm}
\caption{Pretraining the convolutional backbone using momentum contrastive learning improves over previous results by $-4.7\%$ in terms of mesh error, $3.6\%$ in terms of F@\SI{5}{mm} and $0.9\%$ in terms of F@\SI{15}{mm} (comparing Zimmermann \etal\cite{DBLP:conf/iccv/ZimmermannCYRAB19} and \textit{Ours-Multi View}).
This table shows that fixing the \textit{MoCo} learned convolutional backbone and only training the fully-connected part during hand shape estimation (see \textit{Ours-Fixed}) can not compete with state-of-the-art approaches.
\textit{Ours-Fixed-BN} shows that additionally training the batch normalization parameters leads to substantial improvements.
Consequently, leaving all parameters open for optimization (\textit{Ours-Augmentation}) leads to further improvements.
In \textit{Ours-Scratch} all parameters are trained from random initialization, which performs much better than fixing the convolutional layers, but is still behind the reported results in literature, which illustrates the importance of a good network initialization. 
Applying, our proposed sampling strategies \textit{Ours-Background}, \textit{Ours-Temporal} or \textit{Ours-Multi View} does improve results over using an augmentation based sampling strategy like Chen \etal\cite{DBLP:journals/corr/abs-2003-04297}, denoted by \textit{Ours-Augmentation}.
}
\label{tab:quanti_results}
\end{table*}

\noindent\textbf{Hand Shape Estimation}
For comparison we follow \cite{DBLP:conf/iccv/ZimmermannCYRAB19} and rely on the established metrics \emph{mesh error} in \SI{}{cm} and $F$-score evaluated at thresholds of \SI{5}{mm} and \SI{15}{mm}. All of them being reported for the Procrustes aligned estimates as calculated by the online Codalab evaluation service \cite{codalabFreiHand}.

The results of our approach are compared to literature in \tabref{tab:quanti_results}. The results reported share a similar architecture that is based on a ResNet-50 and the differences can be attributed to the training objective and data used.
The approach presented by Boukhayma \etal\cite{DBLP:conf/cvpr/BoukhaymaBT19} performs pretraining on a large synthetic dataset and subsequent training on a combination of datasets containing real images with 2D, 3D and hand segmentation annotation. The datasets used are \textit{MPII+NZSL} \cite{DBLP:conf/cvpr/SimonJMS17}, \textit{Panoptic} \cite{DBLP:conf/cvpr/SimonJMS17} and \textit{Stereo} \cite{DBLP:journals/corr/ZhangJCQXY16}.
Another rendered dataset is proposed and used for training by Hasson \etal\cite{DBLP:conf/cvpr/HassonVTKBLS19}, which is combined with real images from Garcia \etal\cite{DBLP:conf/cvpr/Garcia-Hernando18}.
Zimmermann \etal\cite{DBLP:conf/iccv/ZimmermannCYRAB19} use only the real images from the \textit{FreiHAND} dataset for training, which is the setting we are also using for the results reported.

\tabref{tab:quanti_results} summarizes the results of the quantitative evaluation on \textit{FreiHAND}. It shows that training the network from random initialization leads to unsatisfactory results reported by \textit{Ours-Scratch}, which indicates that a proper network initialization is important. \textit{Ours-Fixed} is training only the fully-connected layers starting from the weights found by \textit{MoCo} while keeping the convolutional part fixed. This achieves results that fall far behind in comparison. Additionally, training the parameters associated with batch normalization gives a significant boost in accuracy as reported by \textit{Ours-Fixed-BN}. The entry named \textit{Ours-Augmentation} does not make use of the advanced background randomization methods, and is the direct application of the proposed \textit{MoCo} approach \cite{DBLP:journals/corr/abs-2003-04297} to our data. In this case all network weights are being trained. It performs significantly better than the fixed approaches and training from scratch, but lacks behind ImageNet based initialization used by Zimmermann \etal\cite{DBLP:conf/iccv/ZimmermannCYRAB19}. Finally, the rows \textit{Ours-Background}, \textit{Ours-Temporal} and \textit{Ours-Multi view} report results for the proposed sampling strategies for positive pairs while training all network parameters. All methods are able to outperform the ImageNet trained baseline by Zimmermann \etal\cite{DBLP:conf/iccv/ZimmermannCYRAB19}. We find the differences between \textit{Ours-Background} and \textit{Ours-Temporal} to be negligible, while \textit{Ours-Multi view} shows an significant improvement over Zimmermann \etal\cite{DBLP:conf/iccv/ZimmermannCYRAB19}. This shows the influence and importance of the proposed sampling strategies. 

\begin{figure}[!tb]
    \centering
    \includegraphics[width=.75\linewidth]{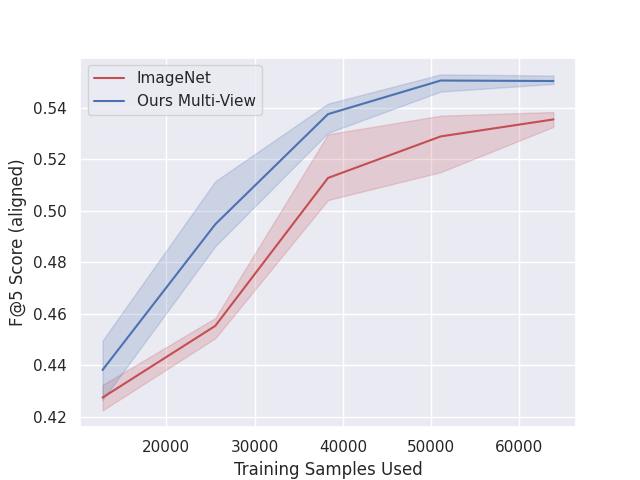}
    \caption{Comparison between our proposed pretraining method (blue) and an ImageNet baseline (red) for varying fractions of the training dataset used during supervised learning of hand shape estimation. The lines represent the mean result over 3 runs per setting, while the shaded area indicates the range of obtained results.
    Our proposed \textit{Multi-View} sampling approach consistently outperforms the baseline, with the largest differences occurring when using approximately 40\% of the full training dataset, 
    indicating a sweet-spot where pretraining is most beneficial. Learning from very large or small datasets reduces the gains from pretraining, for small datasets we hypothesize that there is not sufficient variation in the training dataset to properly learn the task of hand shape estimation anymore.
    }
    \label{fig:db_sparsity}
\end{figure}

To quantify the effect of our proposed multi-view pretraining strategy we do an ablation study in which the quantity of training samples for supervised training is reduced and we evaluate how well our proposed multi-view sampling strategy compares to an ImageNet pretrained baseline. This is shown in \figref{fig:db_sparsity}. In this experiment, the amount of labeled data used during supervised training is varied between 20\% and 100\% of the full \textit{FreiHAND} training dataset, which corresponds to between 12,772 and 63,864 training samples. Except for varying the training data we follow the approach by Zimmermann \etal\cite{DBLP:conf/iccv/ZimmermannCYRAB19}.
The figure shows curves for the minimal, maximal, and mean F@5 scores achieved over three runs per setting for  Procrustes aligned predictions.

We observe that \textit{Our Multi-View} consistently outperforms the ImageNet pretrained baseline and that more data improves the performance of both methods. The differences between both methods are largest when using 40 \% of the full training dataset, for a very large or small supervised datasets the differences between the two methods become smaller. 
We hypothesize, that there is a sweet spot between enough training data, to diminish the value of pretraining, and a minimal amount of training data needed to learn the task of hand shape estimation reasonably well from the supervised data.

A qualitative comparison between our method and results by \cite{DBLP:conf/iccv/ZimmermannCYRAB19} is presented in \figref{fig:qual_results}. It shows that the differences between both methods are visually subtle, but in general hand articulation is captured more accurately by our methods which results into lower \emph{mesh error} and higher $F$-scores.

    \section{Conclusion}
Our work shows that unsupervised visual representation learning is beneficial for hand shape estimation and that sampling meaningful positive pairs is crucial. We only scratch on the surface of possible sampling strategies and find that sampling needs to be inline with the downstream task at hand.

By making the data available publicly, we encourage the community strongly to explore some of the open directions. These include to extend sampling towards other sources of consistency that remain untouched by our work, \eg{}, temporal consistency of hand pose within one recorded sequence leaves opportunities for future exploration.

Another direction points into combining recently proposed weak-supervision methods with the presented pretraining methods and leverage the given calibration information at training time.

%
%
\bibliographystyle{splncs04}
\bibliography{references}

\end{document}